\documentclass{article}

\usepackage{arxiv}
\usepackage{microtype}
\usepackage{graphicx}
\usepackage{booktabs} 

\usepackage[utf8]{inputenc} 
\usepackage[T1]{fontenc}    
\usepackage{url}            
\usepackage{nicefrac}       
\usepackage{microtype} 

\usepackage{caption}
\usepackage{subcaption}
\usepackage{xcolor}
\usepackage[linesnumbered,vlined,ruled,norelsize,algo2e]{algorithm2e}

\usepackage[cmex10]{mathtools} 
\usepackage{amsmath}
\usepackage{dsfont,amssymb,bm}

\usepackage[standard,amsmath,thmmarks] {ntheorem}
\allowdisplaybreaks

\usepackage{flushend}

\theoremstyle{plain}

\usepackage[T1]{fontenc}

\DeclareRobustCommand{\ALPHABET}{\mathcal}

\newcommand{\IND}{\mathds{1}}
\newcommand{\PR}{\mathds{P}}
\newcommand{\EXP}{\mathds{E}}

\newcommand{\reals}{\mathds{R}}

\newcommand{\DEFINED}{\coloneqq}

\title{Option-critic in cooperative multi-agent systems}  

\author{
	Jhelum Chakravorty\textsuperscript{1,3} 
	\And Nadeem Ward\textsuperscript{1,3}
	\And Julien Roy\textsuperscript{2,3}
	\And Maxime Chevalier-Boisvert\textsuperscript{3}\\
	\And Sumana Basu\textsuperscript{1,3}
	\And Andrei Lupu\textsuperscript{1,3}
	\And Doina Precup\textsuperscript{1,3,4}\\
	\and
	\textsuperscript{1}McGill University, \textsuperscript{2}Université de Montréal, 
	 \textsuperscript{3}Mila, \textsuperscript{4}DeepMind,\\
	 \and
	 \texttt{\{jhelum.chakravorty, patrick.ward\}@mail.mcgill.ca,  \{jul.roy1311, maximechevalierb\}@gmail.com}, \\
	 \and \texttt{\{sumana.basu, andrei.lupu\}@mail.mcgill.ca, dprecup@cs.mcgill.ca} \\
}

\begin{document}
\maketitle


\begin{abstract}
In this paper, we investigate learning temporal abstractions in cooperative multi-agent systems, using the options framework (Sutton et al, 1999). First, we address the planning problem for the decentralized POMDP represented by the multi-agent system, by introducing a \emph{common information approach}. We use the notion of \emph{common beliefs} and   broadcasting to solve an equivalent centralized POMDP problem. Then, we propose the Distributed Option Critic (DOC) algorithm, which uses centralized option evaluation and decentralized intra-option improvement. We theoretically analyze the asymptotic convergence of DOC and build a new multi-agent environment to demonstrate its validity. Our experiments empirically show that DOC performs competitively against baselines and scales with the number of agents.
\end{abstract}


\section{Introduction}\label{sec:intro}

\emph{Temporal abstraction} refers to the ability of an intelligent agent to reason, act and plan at multiple time scales~\cite{BartoMahadevan}. A standard way to include temporal abstraction in reinforcement learning agents is through the  \emph{options} framework~\cite{SuttonEtAl1999}.
In~\cite{Bacon2017TheOA}, the authors an approach for learning options, using a gradient-based approach. 

 Multi-agent systems present challenges due to the exacerbated \emph{curse of dimensionality} and \emph{non-classical information structure}. 
Cooperative multi-agent systems or dynamic team problems are decentralized control problems within which the participating agents share rewards and aim to accomplish a common goal, but have access to different information sets (see \cite{mahajan_et_al2012} and references therein for details). The decentralized nature of the information prevents
the use of classical tools in centralized decision theory, such
as dynamic programming, convex analytic methods, or linear
programming. A common formulation of such systems is given by decentralized Markov Decision Processes (Dec-MDPs) and decentralized partially observable Markov Decision Processes (Dec-POMDPs).  Dec-POMDPs offer a very general, sequential, synchronized decision-making framework, but finding the optimal solution for a finite-horizon Dec-POMDP is NEXP-complete, and the infinite-horizon problem is undecidable~\cite{BernsteinEtAL2002}. 
\subsection{Related work}
In this paper we study the option framework for a multi-agent system in  a cooperative setting~\cite{marschak1954,radner1962}. The difficulty pertaining to the combinatorial nature of Dec-POMDP can be mitigated by using the \emph{common information approach}~\cite{MT:NCS}, in which the agents share a common pool of information, which they can use in addition to their own private information; a similar idea was presented recently in~\cite{FoersterEtAlarxiv2018}. However, learning optimal policies in dynamic teams is still quite challenging and updating the common belief in a scalable way is a non-trivial problem. Omidshafiei et al~\cite{OmiEtAL2015,OmiEtAL2016} discuss the problem of solving Decentralized Partially Observable Semi-Markov Decision Processes (Dec-POSMDPs), in which, like in the options framework, single time-step transitions are replaced by actions whose duration is stochastic and conditional on the state and action.  
Makar et al~\cite{makar2001} attack the curse of dimensionality in cooperative multi-agent problems using the MAXQ framework for temporal abstraction~\cite{dietterich2000} but their work requires a hand-designed decomposition of the problem based on prior knowledge, whereas we aim to learn this decomposition from data. Some recent works, e.g.~\cite{vanSeijen2019}, has shown that using multiple agents which are trained on different rewards can help solve large-scale reinforcement learning problems better than a single agent. However, in this case the multiple agents are not really autonomous and they all share the state information (apart from having different rewards), which is not the case in authentic cooperative tasks which consist of fully independent agents. In~\cite{HanEtAl2019}, the authors propose independent Q-learning with  dynamic termination, where each agent estimates decision of joint-policy using delayed broadcast options from all agents. 
In~\cite{Sukhbaatar2016}, multi-agent communication is learnt through backpropagation where the agents networks are interconnected and treated as a big network as a whole, which is not needed in our case.

\subsection{Contribution}
Our contribution in this paper is threefold. 
\begin{itemize}

    \item We formally define the options framework in the cooperative multi-agent setting modelled as a Dec-POMDP. We focus on the information structure in a Dec-POMDP setting and introduce the \emph{common information approach} of forming a \emph{common belief}, which is applicable to both intermittent and continuous communication among agents. We apply this approach in the option framework to find the solution of such a Dec-POMDP. We formulate a suitable dynamic program and establish the optimality of the solution.
    
    \item We propose Distributed Option Critic (DOC), a model-free reinforcement learning algorithm which allows for solving this problem incrementally from data. We analyze the asymptotic convergence of this algorithm. Empirical results also show that DOC is competitive against baselines.
    
    \item We build \textit{TEAMGrid}, a new set of multi-agent gridworld environments based off of Minigrid \cite{gym_minigrid}.

\end{itemize}

\section{Preliminaries}\label{secS:prelims}

We denote vectors by bold script. 
For any set $\ALPHABET C$, $\mathrm{Pow}(\ALPHABET C)$ denotes the power-set of $\ALPHABET C$. We use the shorthand  $x_{1:t}$ to represent the sequence $\{x_1, \dots, x_t\}$. For any space $\ALPHABET X$, $\Delta(\ALPHABET X)$ denotes the space of probability distributions over $\ALPHABET X$. $\ALPHABET S$, $\ALPHABET A$ and $\ALPHABET O$ denote the finite spaces of joint-states, joint-actions and joint-observations os a DEC-POMDP respectively.

As described in ~\cite{Oliehoek:book} the dynamics of the multi-agent system operates in discrete time, as given by:
\begin{equation}\label{eq:dynamics}
\mathbf s_{t+1} = f_t(\mathbf s_t, \mathbf a_t, \mathbf w_t),
\end{equation}
where $f_t$ is a deterministic function dependent on the environment, and $\mathbf s_t$ and $\mathbf a_t$ are the joint-state and the joint-action of the agents at time $t$; $\mathbf w_t$ is the system noise vector represented by a stochastic process. 

The \emph{value function} measures the performance of a Dec-POMDP, which is the expected reward over the finite or infinite time horizon, where the reward is acieved by a joint-policy. The expectation depends on the joint transition probability which is completely specified by the transition and observation model and the joint
policy~\cite{olihoek_dec_Q}. In case of \emph{infinite horizon discounted reward}, which is our case, the value function measures the expected discounted reward over infinite horizon. In this paper we assume bounded per-step reward and each agent's reward depends only on its current state, current action and next state (\emph{reward independent} agents).

In a Dec-POMDP, the agents do not have complete knowledge of others' states (and sometimes even their own states); instead, they share a \emph{common information} which they update by communicating at every step (\emph{cheap talk} or always broadcasting) or intermittently (\emph{intermittent broadcasting}). In the cooperative setting, a centralized value function (or critic) evaluates the performance of the agents. 
In this paper, we consider both communications.
$r^{\mathbf a_t}(\mathbf s_t)$ is the immediate reward of choosing action $\mathbf a_t$ in state $\mathbf s_t$
For reward independent Dec-POMDPs, such as ours, $r^{\mathbf a_t}(\mathbf s_t) = \sum_{j \in \ALPHABET J} R^j(s^j_t, a^j_t, s^j_{t+1})$, $p^{\mathbf a_t}(\mathbf s_t,\mathbf s_{t+1})$ is the one-step transition probability from joint-state $\mathbf s_t$ to $\mathbf s_{t+1}$ under joint-action $\mathbf a_t$.
 $\gamma \in (0,1)$ is the \emph{discount factor}.
 
\section{Temporal abstraction with full observability}\label{subsecS:temporal-abstraction}

In this paper, we consider \emph{Markov} options which execute in \emph{call-and-return} way; we will now define these notions in the context of a multi-agent system (see~\cite{SuttonEtAl1999} for more details). 

In a fully observable multi-agent environment with $J$ agents, a \emph{Markov} joint-option $\boldsymbol{\omega}$ consists of a vector of component options for each agent, $\boldsymbol{\omega} = (\omega^1,\dots,\omega^J)$. It can initiate, if no other option is currently executing, at joint-state which is part of its initiation set $\mathbf s \in I^{\boldsymbol{\omega}}$. If $\boldsymbol{\omega}$ is executing at time $t$, it generates joint-action $\mathbf a_t$ according to $a^j_t \sim \pi^{\omega^j_t}_t(\cdot|s^j_t)$. The environment then generates next joint-state $\mathbf s_{t+1}$, where the option $\omega^j_t$ terminates with probability $\beta^{\omega^j_t}_t(s^j_{t+1})$, $\beta^{\omega^j_t}_t(s^j_{t+1}) \in (0,1]$. If any of the component options terminates, then the joint option also terminates and a new joint-option has to be chosen. Otherwise, the joint-action selection process continues as above. We will denote by $\mu$  the policy which chooses joint-options.

Let $\ALPHABET E(\omega^j_t,s^j_t)$, $j \in \ALPHABET J$ be the event that  $\omega^j$ is initiated at state $s^j$ at time $t$. Let $m$ be a random variable indicating the time elapsed since $t$. Then, the reward of Agent~$j$, $r^{\omega^j_t}(s^j_t)$ until termination of $\omega^j_t$ is:
\begin{equation}\label{eq:option-reward}
  r^{\omega^j_t}(s^j_t) \DEFINED \EXP \Big[\sum_{\tau = t}^{t+m} \gamma^{\tau-t} R^j(s^j_\tau, a^j_\tau, s^{j}_{\tau+1})\,|\, \ALPHABET E(\omega^j_t, s^j_t)\Big],
\end{equation}
where $\EXP[\cdot]$ denotes expectation, $R^j$ is the  reward of agent~$j$, actions $a^j_\tau$ are generated according to the internal policy $\pi^{\omega^j_t}_t$ of option $\omega^j_t$.
For ease of exposition, we write $r^j_{t+1}=R^j(s^j_t, a^j_{t}, s^{j}_{t+1})$.
Note that~\eqref{eq:option-reward} can be expanded recursively as follows:
\[
 r^{\omega^j_t}(s^j_t) = \beta^{\omega^j_t}_t(s^j_{t+1})
 r^j_{t+1} + \gamma (1-\beta^{\omega^j_t}_t(s^j_{t+1})) r^{\omega^j_t}(s^j_{t+1}),
\]
The total reward $r^{\boldsymbol{\omega}_t}(\mathbf s_t)$  for joint option $\boldsymbol{\omega}_t = (\omega^1_t,\dots, \omega^J_t)$ is given by
\begin{equation}\label{eq:total-joint-option-reward}
r^{\boldsymbol{\omega}_t}(\mathbf s_t) \DEFINED \sum_{j \in \ALPHABET J} r^{\omega^j_t}(s^j_t).
\end{equation}

Next, let $p^{\boldsymbol{\omega}_t}(s,s')$ denote the probability of choosing joint-option $\boldsymbol{\omega}_t$ at state $s$ and transitioning to state $s'$, where $\boldsymbol{\omega}_t$ terminates, i.e., $p^{\boldsymbol{\omega}_t}(\mathbf s, \mathbf s') \DEFINED \PR(\mathbf s_{t'} = \mathbf s'\,|\, \ALPHABET E(\boldsymbol{\omega}_t, \mathbf s_t = \mathbf s))$ for any $t'>t$. Then
\begin{equation}\label{eq:option-transition}
p^{\boldsymbol{\omega}_t}(\mathbf s, \mathbf s') \DEFINED \sum_{m = 1}^\infty p^{\boldsymbol{\omega}_t}_m(\mathbf s, \mathbf s'),
\end{equation}
where $p^{\boldsymbol{\omega}_t}_m(\mathbf s, \mathbf s')$ is the probability that a joint-option $\boldsymbol{\omega}_t$ initiated in joint-state $\mathbf s$ at time $t$ terminates in joint-state $\mathbf s'$ after $m$ steps. 

Let $\beta^{\boldsymbol{\omega}_t}_{\text{none}}(\mathbf s_t)$ be the probability of no agent terminating in joint-state $\mathbf s_t$. 
From the independence of agents we have:
\begin{equation}\label{eq:beta_all}
    \beta^{\boldsymbol{\omega}_t}_{\text{none}}(\mathbf s_t) = \prod_{j \in \ALPHABET J} (1-\beta^{\omega^j_t}_t(s^j_t)).
\end{equation} 
Then, $p^{\boldsymbol{\omega}_t}_m(\mathbf s, \mathbf s')$ can be expanded recursively as follows:
\[
p^{\boldsymbol{\omega}_t}_m(\mathbf s, \mathbf s') 
 = \gamma \sum_{a^j \in \ALPHABET A^j} \Big[\pi^{\boldsymbol{\omega}_t}_t(\mathbf a_t = \mathbf a|\mathbf s_t=\mathbf s)
\sum_{\mathbf s'' \in \ALPHABET S}\PR(\mathbf s_{t+1} = \mathbf s''|\mathbf s_t=\mathbf s, \mathbf a_t = \mathbf a) \beta^{\boldsymbol{\omega}_t}_{\text{none}}(\mathbf s)p^{\boldsymbol{\omega}_t}_{m-1}(\mathbf s'', \mathbf s')\Big].
\]
Let $\ALPHABET M$ be the space of Markov option-policies $\mu_t: \ALPHABET S \rightarrow \Delta(\Omega)$. We denote $\mu_t(\boldsymbol{\omega}_t|\mathbf s)=\mu_t(\boldsymbol{\omega}_t|\mathbf s_t = \mathbf s)$.
Following~\cite{SuttonEtAl1999}, let $U^\mu_t(\mathbf s_t,\boldsymbol{\omega}_t)$ be the option-value \emph{upon arrival} at joint-state $\mathbf s_t$ using option-policy $\mu_t$:
\begin{equation}
U^{\mu_t}(\mathbf s_t, \boldsymbol{\omega}_t) 
\DEFINED 
 \beta^{\boldsymbol{\omega}_t}_{\text{none}}(\mathbf s_t)  Q^{\mu_t}(\mathbf s_t, \boldsymbol{\omega}_t)
   + (1-\beta^{\boldsymbol{\omega}_t}_{\text{none}}(\mathbf s_t)) \max_{\ALPHABET T \in \mathrm{Pow}(\ALPHABET J)}\max_{\boldsymbol{\omega'}_t \in \Omega(\ALPHABET T)} Q^{\mu_t}(\mathbf s_t, \boldsymbol{\omega'}_t), \label{eq:U-mu}
\end{equation}
where we use a slight abuse of notation, $\boldsymbol{\omega'}_t$, to mean $\boldsymbol{\omega}_t = \boldsymbol{\omega'}$, $\Omega(\ALPHABET T)$ denotes the set of options for agents in $\ALPHABET T \subseteq \ALPHABET J$, where $\ALPHABET T$ is the set of the agents terminating their current options.

$Q^{\mu_t}$ in~\eqref{eq:U-mu} is the solution of the following Bellman update:
\begin{equation}
Q^{\mu_t}(\mathbf s_t, \boldsymbol{\omega}_t) 
= \sum_{\mathbf a_t \in \ALPHABET A} \pi^{\boldsymbol{\omega}_t}_t(\mathbf a_t|\mathbf s_t) \Big[r^{\mathbf a_t}(\mathbf s_t) 
 + \gamma \sum_{\mathbf s_{t+1} \in \ALPHABET S}\big(p^{\mathbf a_t}(\mathbf s_t,\mathbf s_{t+1}) U^{\mu_t}(\mathbf{s_{t+1}}, \boldsymbol{\omega}_t) \big)\Big], \label{eq:Q-mu}
\end{equation}
where $\pi^{\boldsymbol{\omega}_t}_t(\mathbf a_t|\mathbf s_t)$\footnote{For agents with factored actions such as ours, $\pi^{\boldsymbol{\omega}_t}_t(\mathbf a_t|\mathbf s_t) = \prod_{j} \pi^{\omega^j_t}_t(a^j_t|s^j_t)$.} is the shorthand for the action-policy to choose joint-action $\mathbf a_t$ under joint-option $\boldsymbol{\omega}_t$ in joint-state $\mathbf s_t$. 
We denote by  $U^*$ and  $Q^*$ the corresponding optimal values.

The \emph{dynamic team problem} that we are interested to solve is to choose policies that maximize the the infinite-horizon discounted reward:
$\ALPHABET R^{\mu_t}$ as given by
\begin{equation}
\sup_{\mu_t \in \mathcal M}  
\sum_{\boldsymbol{\omega}_t \in \Omega} 
\mu_t(\boldsymbol{\omega}_t\, | \,\mathbf s_t) 
\EXP{E} \Bigg[\sum_{t = 0}^\infty \gamma^t
  r_{t+1}\,|\, \ALPHABET E(\boldsymbol{\omega}_0 \mu_0,\mathbf s_0)\Bigg], \label{eq:R-problem1}
\end{equation}

\section{Dec-POMDP planning with temporal abstraction}\label{sec:planning}
The Common Information Approach~\cite{NMT:partial-history-sharing} is an effective way to solve a Dec-POMDP in which the agents share a common pool of information, updated eg via boradcasting,
in addition to \emph{private} information available only to each individual agent. A \emph{fictitious coordinator} observes the common information and suggests a \emph{prescription} - in our case the Markov joint-option policy $\mu_t$). The joint-option $\boldsymbol{\omega}_t$ is chosen from $\mu_t$ and is communicated to all agents~$j$, who in turn generate their own action $a^j_t$ according to their local (private) information, and their own observation $o^j_t$ : $a^j_t \sim \pi^j_t(a^j_t|o^j_t)$. A 
\emph{locally fully observable} agent chooses its action $a^j_t$ based on its own state $s^j_t$ or embedding $e^j_t$ 
according to $a^j_t \sim \pi^j_t(a^j_t|s^j_t)$\footnote{For ease of exposition we use the notation for states but the same analysis applies to the embeddings.}
The notion of a centralized fictitious coordinator transforms the Dec-POMDP into an equivalent centralized POMDP, so one can exploit mathematical tools from stochastic optimization such as  dynamic programming to find an optimal solution. 

The common information-based belief on the joint-state $\mathbf s_t \in \ALPHABET S$ is defined as:
\begin{equation}\label{eq:belief-option}
b^c_t(\mathbf s) \DEFINED \PR(\mathbf s_t = \mathbf s\,|\, \ALPHABET I^c_t),
\end{equation}
where $\ALPHABET{I}^c_t$ is the common information at time  $t$.

Let $\mathrm{Broad}(o^j_t,\omega^j_t) = \mathrm{br}^j \in \{0,1\}$ be the broadcast symbol, where $\mathrm{br}^j = 1$ if Agent~$j$ has broadcast and $0$ otherwise\footnote{In general there can be finite number of levels of broadcast, instead of binary levels. In this paper we use binary levels since that is sufficient for our purpose but the results are extendable to finite number of levels.}. When Agent~$j$ decides to broadcast, its observation $o^j_t$ is received by all other agents. Hence, the common information is $\boldsymbol{\tilde o}_t = (\tilde o^1_t, \dots, \tilde o^J_t)$, where $\tilde o^j_t$, $j \in \ALPHABET J$ is given by
\begin{equation}\label{eq:tilde-o}
\tilde o^j_t\DEFINED \begin{cases}
                  o^j_t &\mbox{if $\mathrm{Broad}(o^j_t,\omega^j_t) = 1$}\\
                  \varnothing, &\mbox{otherwise}.
                  \end{cases}
\end{equation}
The coordinator observes $\boldsymbol{\tilde o}_{1:t}$, and generates $\mu_t$, according to some \emph{coordination rule} $\psi$ such that $\psi: (\ALPHABET O \cup \{\varnothing\})^{t-1} \rightarrow \ALPHABET M$, $j \in \ALPHABET J$, 
\begin{equation}
  \mu_t = \psi(\boldsymbol{\tilde o}_{1:t-1}, \mu_{1:t-1}) \label{eq:mu},
\end{equation}
The options $\omega^j_t \in \boldsymbol{\omega}_t$, $\boldsymbol{\omega}_t \sim \mu_t$, are then communicated to all agents. Thus, $\ALPHABET I^c_t$ appearing in~\eqref{eq:belief-option} is given by:
\[
\ALPHABET I^c_t = \{\boldsymbol{\tilde o}_{1:t-1}, \boldsymbol{\omega}_{1:t-1}\},
\]
and thus, $\ALPHABET I^c_{t-1} \subseteq \ALPHABET I^c_t$. Consequently,~\eqref{eq:belief-option} can be rewritten as:
\begin{equation}\label{eq:belief-option-2}
b^c_t(\mathbf s) \DEFINED \PR(\mathbf s_t = \mathbf s\,|\, \boldsymbol{\tilde o}_{1:t-1}, \boldsymbol{\omega}_{1:t-1}).
\end{equation}
Upon receiving $\omega^j_t$, Agent~$j$ uses the action-policy $\pi^{\omega^j_t}_t$ and termination probability $\beta^{\omega^j_t}_t$ corresponding to $\omega^j_t$ and generates its action $a^j_t$ using its local information $o^j_t$ as per $a^j_t \sim \pi^{\omega^j_t}_t(a^j_t|o^j_t)$. 

From~\eqref{eq:belief-option-2}, $b^c_t$ is measurable with $(\boldsymbol{\tilde o}_{1:t-1}, \mu_{1:t-1})$, so also using~\eqref{eq:mu}, we can infer
that there is no loss of optimality if we restrict attention to coordination rules $\tilde \psi$ such that:
\begin{equation}\label{eq:mu_t_psi_tilde}
  \mu_t = \tilde \psi(b^c_t).
\end{equation}
The posterior of the common information based belief $b^c_t$ can then be written as
\begin{equation}\label{eq:belief-update}
  b^c_{t,t} = h_t(b^c_t, \boldsymbol{\tilde o}_t, \mu_t),
\end{equation}
where $b^c_{t,t}(\mathbf s) \DEFINED \PR(\mathbf s_t = \mathbf s\,|\,\boldsymbol{\tilde o}_{1:t}, \boldsymbol{\omega}_{1:t})$ and the function $h_t$ is the Bayesian filtering update function\footnote{Bayesian filtering applies Bayesian statistics and Bayes' rule in solving Bayesian inference problems including stochastic filtering problems. Iterative Bayesian learning was introduced by~\cite{HoLee1964} (among others), which involves Kalman filtering as a special case. See~\cite{Chen2003} and references therein for details.}. Consequently, we have
\begin{align}\label{eq:belief-update-t+1}
    & \hskip -1em b^c_{t+1}(\mathbf s') \DEFINED \PR(\mathbf s_{t+1} = \mathbf s'\,|\,\boldsymbol{\tilde o}_{1:t}, \boldsymbol{\omega}_{1:t}) 
    = \sum_{\mathbf s \in \ALPHABET S} p^{\mathbf a_t}(\mathbf s, \mathbf s')b^c_{t,t}(\mathbf s).
\end{align}

Using the argument  of~\cite[Lemma~1]{NMT:partial-history-sharing}, we can show that the coordinated system is a POMDP with prescriptions $\mu_t$ and observations 
\begin{equation}\label{eq:o-tildeh}
  \boldsymbol{\tilde o}_t = \tilde h_t(  \mathbf s_t, \mu_t),
\end{equation}
Furthermore, define $ \mathbf o^\dagger_t \DEFINED \boldsymbol{\tilde o}_{1:t-1}$. Then:
\begin{equation}\label{eq:prob_o_coordinated_sys}
  \PR(\mathbf o^\dagger_{t+1}= \mathbf o^\dagger\,|\,  \mathbf o^\dagger_{1:t}, \mu_{1:t}) = \PR( \mathbf o^\dagger_{t+1}=\mathbf o^\dagger\,|\,  \mathbf o^\dagger_t, \mu_t),
\end{equation}
where  $\mathbf o^\dagger$ denotes the realization of the sequence $\boldsymbol{\tilde o}_{1:t}$, which behaves like a state.

This relies on showing equalities of conditional probability values by shedding off \emph{irrelevant information}. Note that while computing the conditional probability in~\eqref{eq:prob_o_coordinated_sys}, the information captured in $\mathbf o^\dagger_{1:t}$ and $\boldsymbol{\tilde o}_{1:t-1} \eqqcolon \mathbf o^\dagger_t$ are the same. So, $\mathbf o^\dagger_{1:t-1}$ can be considered redundant (and thus irrelevant) information and can hence be removed from conditioning. The common-observation $\boldsymbol{\tilde o}_t$ depends on the joint-state $\mathbf s_t$ and the joint option-policy $\mu_t$ (through $\tilde h_t$). So, when conditioned by $\mu_t$, $\mu_{1:t-1}$ does not give any additional information about $\boldsymbol{\tilde o}_t$ and can thus be removed from conditioning as well.

The Bayesian update for the posterior, $b^c_{t,t}$ is:
\begin{equation}\label{eq:belief-Bayes-update-options}
  b^c_{t,t} 
  = \begin{cases}
                    \mathrm{DIRAC}(\mathbf o_t), & \mbox{if $\boldsymbol{\tilde o}_t \neq \varnothing$}\\
                 \alpha_{b^c_t, \boldsymbol{\tilde o}_t},  &\mbox{otherwise},
           \end{cases}
\end{equation}
where by $\boldsymbol{\tilde o}_t \neq \varnothing$ we mean that all agents have broadcast, $\mathrm{DIRAC}(\mathbf o_t)$ is the Dirac-delta distribution at $\mathbf o_t$. The function $\alpha_{b^c_t, \boldsymbol{\tilde o}_t}$ is given by:
\[
\alpha_{b^c_t, \mathbf \boldsymbol{\tilde o}_t}(\mathbf s_t) 
\DEFINED  \frac{\IND(\tilde h_t(\mathbf s_t, \mu_t) = \boldsymbol{\tilde o}_t) b^c_t(\mathbf s_t)}{\sum_{\mathbf s'_t \in \ALPHABET S} \IND(\tilde h_t(\mathbf s'_t, \mu_t) = \boldsymbol{\tilde o}_t) b^c_t(\mathbf s'_t)}, 
\]
where for an event $E$, $\IND(E)$ denotes its indicator function and we use $\mathbf s'_t$ to mean $\mathbf s_t = \mathbf s'$.

Recall the broadcast symbol of Agent~$j$, $\mathrm{br}^j_t \in \{0,1\}$. Then,  $\tilde h_t$ is given by:
\begin{equation}\label{eq:tilde-h}
\tilde h_t(\mathbf s_t, \mu_t) \DEFINED \EXP^{\mu_t}[\PR(\boldsymbol{\tilde o}_t|\mathbf s_t, b^c_t, \boldsymbol{\omega}_t)],
\end{equation}
where
\begin{align}
\PR(\boldsymbol{\tilde o_t}|\mathbf s_t, b^c_t, \boldsymbol{\omega}_t) 
&=\sum_{
 \boldsymbol{\mathrm{br}}\in \{0,1\}^J, \mathbf a_t \in \ALPHABET A}
 \sum_{\mathbf o_t \in \ALPHABET O}
\pi^{b,\boldsymbol{\omega}_t}_t(\boldsymbol{{\mathrm{br}}}_t|\mathbf o_t)\pi^{\boldsymbol{\omega}_t}_t(\mathbf a_t| \mathbf o_t)
f_t(\mathbf o_t, \mathbf s_t, \boldsymbol{\omega}_{t-1}) b^c_t(\mathbf s_t), \label{eq:prob-o-tilde}\\
f_t(\mathbf o_t, \mathbf s_t, \boldsymbol{\omega}_{t-1})
&\DEFINED \sum_{a_{t-1} \in \ALPHABET A} \eta(\mathbf o_t|, \mathbf s_t, \mathbf a_{t-1})\pi^{\boldsymbol{\omega}_{t-1}}_{t-1}(\mathbf a_{t-1}|\mathbf o_{t-1}) 
f_{t-1}(\mathbf o_{t-1}, \mathbf s_{t-1}, \boldsymbol{\omega}_{t-2}) \label{eq:f_t-recursive}.
\end{align}
In~\eqref{eq:prob-o-tilde}, $\pi^{b,\boldsymbol{\omega}_t}_t$ is the joint broadcast-policy and  in~\eqref{eq:f_t-recursive}, $\eta$ is the probability of getting joint-observation $\mathbf o_t$ at a joint-state $\mathbf s_t$, reached by using action $\mathbf a_{t-1}$. For factored agents we have 
\[
\pi^{b,\boldsymbol{\omega}_t}_t(\boldsymbol{\mathrm{br}}_t|\mathbf o_t) = \prod_{j \in \ALPHABET J} \pi^{b, \omega^j_t}_t(\mathrm{br}^j_t|o^j_t), \quad
\eta(\mathbf o_t|\mathbf s_t, \mathbf a_{t-1}) = \prod_{j \in \ALPHABET J} \eta^j(o^j_t|s^j_t, a^j_{t-1}).
\]

The optimal policy of the coordinated centralized system is the solution of a suitable dynamic program which has a fixed-point. In order to formulate this program, we need to show that $b^c_t$ is an \emph{information state}, i.e. a sufficient statistic to form, with the current joint-option $\mu_t$,  a future belief $b_{t+1}^c$. In other words:
\begin{lemma}\label{lemma:info-state-option}
The common information based belief state $b^c_t$ is an information state. In particular,
\begin{enumerate}
  \item $\PR(\mathbf s_t \,|\, \boldsymbol{\tilde o}_{1:t-1}, \boldsymbol{\omega}_{1:t-1}) = \PR(\mathbf s_t\,|\, b^c_t)$
  \item $\PR(b^c_{t+1}\,|\, \boldsymbol{\tilde o}_{1:t-1}, \boldsymbol{\omega}_{1:t-1}) = \PR(b^c_{t+1}\,|\, b^c_t)$
  \item $\EXP [r^{\boldsymbol{\omega}_t}(\mathbf s_t)\,|\, \boldsymbol{\tilde o}_{1:t-1}, \boldsymbol{\omega}_{1:t}] = \EXP [r^{\boldsymbol{\omega}_t}(\mathbf s_t)\,|\, b^c_t, \boldsymbol{\omega}_t]$,
\end{enumerate}
where $r^{\boldsymbol{\omega}_t}$ is given by~\eqref{eq:r-omega}.
\end{lemma}
\begin{proof}
The proof follows an argument similar to~\cite{KumarVaraiya:1986} for primitive actions. In particular,
\begin{enumerate}
  \item The equality of Part~1) readily holds from the fact that $b^c_t$ is measurable by $\boldsymbol{\tilde o}_{1:t-1}$ and so conditioning by $\boldsymbol{\tilde o}_{1:t-1}$ is the same as conditioning by $b^c_t$.
  \item From~\eqref{eq:belief-Bayes-update-options} we can write $\PR(b^c_{t+1}\,|\, \boldsymbol{\tilde o}_{1:t-1}) = \PR(\mathbf s_{t+1}\,|\, \boldsymbol{\bar o}_{1:t-1})\IND(\boldsymbol{\bar o}_{1:t-1} = \boldsymbol{\tilde o}_{1:t-1})$. Then the equality follows from the fact that $b^c_t$ is measurable by $\boldsymbol{\tilde o}_{1:t-1}$ and that conditioning on $b^c_t$ is same as conditioning on $\boldsymbol{\tilde o}_{1:t-1}$ (as is shown by part~1).
  \item We have by the definition of option
  \begin{align*}
  &\EXP [r^{\boldsymbol{\omega}}(\mathbf s)\,|\, \boldsymbol{\tilde o}_{1:t-1}, \boldsymbol{\omega}] = \sum_{a \in \ALPHABET A} \pi^{\boldsymbol{\omega}}(\mathbf a|\mathbf s) \EXP[r^{\mathbf a}(\mathbf s)\,|\,\boldsymbol{\tilde o}_{1:t-1}]\\
  &= \sum_{\mathbf a \in \ALPHABET A} \pi^{\boldsymbol{\omega}}(\mathbf a|\mathbf s) \sum_{\mathbf s \in \ALPHABET S} r^{\mathbf a}(\mathbf s) \PR(\mathbf s_t = \mathbf s \,|\,\boldsymbol{\tilde o}_{1:t-1})
  \stackrel{(a)}{=} \EXP [r^{\boldsymbol{\omega}}(\mathbf s)\,|\, b^c_t, \boldsymbol{\omega}],
  \end{align*}
  where $(a)$ holds by the definition of $b^c_t$.
  \end{enumerate}
  This completes the proof of the lemma. 
\end{proof}

For large systems, the common belief is intractable due to the combinatorial nature of joint state-space. One way to circumvent the combinatorial effect is to assume that the common belief is \emph{factored}~\cite{FoersterEtAlarxiv2018}, i.e., 
\begin{equation}
  b^c_t(\mathbf s) \DEFINED \PR(\mathbf s_t = \mathbf s\,|\, \boldsymbol{\tilde o}_{1:t-1}) \approx \prod_{j \in \ALPHABET J} \PR(s^j_t = s^j\,|\, \boldsymbol{\tilde o}_{1:t-1}) 
  \eqqcolon \prod_{j \in \ALPHABET J} b^{c,\text{fact}}_t(s^j)  \eqqcolon  b^{c,\text{fact}}_t(\mathbf s). \label{eq:factored_common_belief}
\end{equation}
Note that in situations where collision among agents is allowed, common belief becomes factored.
\subsection{Common-belief based option-value}\label{subsec:CI-option-value}
We can extend the notion of option-value with full observability, given by~\eqref{eq:U-mu} and~\eqref{eq:Q-mu} to the case with partial observability. 
The \emph{option-value upon arrival}, $U^\mu$, and the \emph{option-value}, $Q^\mu$, are defined below:
\begin{align}
U^{\mu_t}(b^c_t,\boldsymbol{\omega}_t) &\DEFINED \sum_{\mathbf s_t \in \ALPHABET S} U^{\mu_t}(\mathbf s_t, \boldsymbol{\omega}_t)b^c_t(\mathbf s_t) \notag \\
& =  \sum_{\mathbf s_t \in \ALPHABET S} \Big[\beta^{\boldsymbol{\omega}_t}_{\text{none}}(\mathbf s_t)  Q^{\mu_t}(\mathbf s_t, \boldsymbol{\omega}_t)b^c_t(\mathbf s_t) 
 + (1-\beta^{\boldsymbol{\omega}_t}_{\text{none}}(\mathbf s_t)) \max_{\ALPHABET T \in \mathrm{Pow}(\ALPHABET J)}
\max_{\boldsymbol{\omega'}_t \in \Omega(\ALPHABET T)} Q^\mu(\mathbf s_t, \boldsymbol{\omega}_t')b^c_t(\mathbf s_t)\Big]. \label{eq:U-mu-b}
\end{align}

$Q^{\mu_t}$ in~\eqref{eq:U-mu-b} is the solution of the following Bellman update:
\begin{align}
Q^{\mu_t}(b^c_t,\boldsymbol{\omega}_t) &\DEFINED \sum_{s_t \in \ALPHABET S} Q^{\mu_t}(\mathbf s_t,\boldsymbol{\omega}_t)b^c_t(\mathbf s_t) \notag \\
& = \sum_{\mathbf s_t \in \ALPHABET S}
\sum_{\mathbf o_t \in \ALPHABET O}\Bigg(\sum_{\boldsymbol{\mathrm{br}}_t \in \{0,1\}^J}\sum_{\mathbf a_t \in \ALPHABET A} \pi^{b,\boldsymbol{\omega}_t}_t(\boldsymbol{\mathrm{br}}_t|\mathbf o_t)\pi^{\boldsymbol{\omega}_t}_t(\mathbf a_t| \mathbf s_t) f_t(\mathbf o_t, \mathbf s_t, \boldsymbol{\omega}_{t-1}) \Big[r^{\mathbf a_t, \boldsymbol{\mathrm{br}}_t}(\mathbf s_t)\notag \\
 & \hskip 8em + \gamma\sum_{\mathbf s_{t+1} \in \ALPHABET S} b^c_{t+1}(\mathbf s_{t+1}) \big(p^{\mathbf a_t}(\mathbf s_t,\mathbf s_{t+1}) U^\mu(\mathbf s',\boldsymbol{\omega}_t) \big)\Big] \Bigg)b^c_t(\mathbf s_t), \label{eq:Q-mu-b}
\end{align}
where $f_t(\mathbf o_t, \mathbf s_t, \boldsymbol{\omega}_{t-1})$ is given by~\eqref{eq:f_t-recursive} and $r^{\mathbf a_t, \boldsymbol{\mathrm{br}}_t}(\mathbf s_t)$ is the immediate reward of choosing action $\mathbf a_t$ and broadcast symbol $\boldsymbol{\mathrm{br}}_t$ in state $\mathbf s_t$.
The optimal values corresponding to~\eqref{eq:U-mu-b} and~\eqref{eq:Q-mu-b} are defined as usual.

Define operators $\ALPHABET B^{\mu_t}$ and $\ALPHABET B^*$ as follows:
\begin{align*}
[\ALPHABET B^{\mu_t} Q^{\mu_t}](b^c_t, \boldsymbol{\omega}_t) 
   &\DEFINED \gamma
   \sum_{\mathbf s_t \in \ALPHABET S}
   \sum_{\mathbf o_t \in \ALPHABET O}
   \Bigg(\sum_{\boldsymbol{\mathrm{br}}_t \in \{0,1\}^J}\sum_{\mathbf a_t \in \ALPHABET A}
   \pi^{b,\boldsymbol{\omega}_t}_t(\boldsymbol{\mathrm{br}}_t|\mathbf o_t)\pi^{\boldsymbol{\omega}_t}_t(\mathbf a_t| \mathbf o_t)\\
   & \hskip 8em f_t(\mathbf o_t, \mathbf s_t, \boldsymbol{\omega}_{t-1}) \sum_{\mathbf s_{t+1} \in \ALPHABET S} b^c_{t+1}(\mathbf s_{t+1}) \big(p^{\mathbf a_t}_t(\mathbf s_t, \mathbf s_{t+1}) U^{\mu_t}(\mathbf s_{t+1}, \boldsymbol{\omega}_t) \big) \Bigg)b^c_t(\mathbf s_t),\\
  [\ALPHABET B^* Q^*](b^c_t, \boldsymbol{\omega}_t) 
   &\DEFINED \gamma
   \sum_{\mathbf s_t \in \ALPHABET S}\sum_{\mathbf o_t \in \ALPHABET O}
   \Bigg(\sum_{\boldsymbol{\mathrm{br}}_t \in \{0,1\}^J}\sum_{\mathbf a \in \ALPHABET A}
   \pi^{b,\boldsymbol{\omega}_t}_t(\boldsymbol{\mathrm{br}}_t|\mathbf o_t)\pi^{\boldsymbol{\omega}_t}_t(\mathbf a_t| \mathbf s_t)\\
     & \hskip 8em f_t(\mathbf o_t, \mathbf s_t, \boldsymbol{\omega}_{t-1})  \sum_{\mathbf s_{t+1} \in \ALPHABET S} b^c_{t+1}(\mathbf s_{t+1}) \big(p^{\mathbf a_t}(\mathbf s_t, \mathbf s_{t+1})   \max_{\boldsymbol{\omega'}_t \in \Omega} U^{\mu_t}(\mathbf s_{t+1}, \boldsymbol{\omega'}_t) \big) \Bigg)b^c_t(\mathbf s_t).
\end{align*}
Then, $Q^{\mu_t}$ and $Q^*$ can be rewritten as 
\begin{equation}
Q^{\mu_t}(b^c_t,\boldsymbol{\omega}_t)  = r^{\boldsymbol{\omega}_t}(b^c_t) + [\ALPHABET B^{\mu_t} Q^{\mu_t}](b^c_t,\boldsymbol{\omega}_t), \quad
  Q^*(b^c_t,\boldsymbol{\omega}_t)  = r^{\boldsymbol{\omega}_t}(b^c_t) + [\ALPHABET B^* Q^*](b^c_t,\boldsymbol{\omega}_t), \label{eq:Q-star-2}
\end{equation}
where
\begin{equation}
r^{\boldsymbol{\omega}_t}(b^c_t) \DEFINED \sum_{\mathbf s_t \in \ALPHABET S}
\sum_{\mathbf o_t \in \ALPHABET O}\sum_{\boldsymbol{\mathrm{br}}_t \in \{0,1\}^J} \sum_{\mathbf a_t \in \ALPHABET A}
\pi^{b,\boldsymbol{\omega}_t}_t(\boldsymbol{\mathrm{br}}_t|\mathbf o_t)\pi^{\boldsymbol{\omega}_t}_t(\mathbf a_t| \mathbf o_t)
r^{\mathbf a_t, \boldsymbol{\mathrm{br}}_t}(\mathbf s_t)  f_t(\mathbf o_t, \mathbf s_t, \boldsymbol{\omega}_{t-1}) b^c_t(\mathbf s_t) \label{eq:r-omega}.
\end{equation}

\begin{lemma}\label{lemma:Q-star-contraction}
The operators $\ALPHABET B^*$ and $\ALPHABET B^{\mu_t}$ are contractions. In particular, for any $\gamma \in (0,1)$,
\[
  \|\ALPHABET B^{\mu_t} Q^{\mu_t}\|_\infty \le \gamma \|Q^{\mu_t}\|_\infty,
  \quad
  \|\ALPHABET B^* Q^*\|_\infty \le \gamma \|Q^*\|_\infty
\]
where $\|\cdot\|_\infty$ is the sup-norm. 
\end{lemma}
\begin{proof}
We prove the contraction of $\ALPHABET B^*$. That $\ALPHABET B^\mu$ is a contraction can be shown similarly. 

We begin by noting that the supremum in the definition of the sup-norm can be replaced by maximum since $\ALPHABET S$ is finite. Then, we have 
\begin{align*}
&\|\ALPHABET B^* Q^*\|_\infty \\
&= \gamma \max_{b^c_t \in \Delta(\ALPHABET S)}\max_{\boldsymbol{\omega} \in \Omega} \sum_{\mathbf s \in \ALPHABET S}\Big[ \sum_{\mathbf a \in \ALPHABET A} \pi^{\boldsymbol{\omega}}(\mathbf a | \mathbf s)\\
& \hskip 2em \times 
 \Big( \sum_{\mathbf s' \in \ALPHABET S} b^c_{t+1}(\mathbf s') p^{\mathbf a}(\mathbf s,\mathbf s') U^*(\mathbf s', \mathbf o)\Big)\Big]b^c_t(\mathbf s)\\
& \stackrel{(a)}{\le} \gamma \max_{b^c_t \in \Delta(\ALPHABET S)}\|Q^*\|_\infty \sum_{\mathbf s \in \ALPHABET S} \Big[ \sum_{\mathbf a \in \ALPHABET A} \pi^{\boldsymbol{\omega}}(\mathbf a | \mathbf s)\\
 & \hskip 2em \times \Big( \sum_{\mathbf s' \in \ALPHABET S} b^c_{t+1}(\mathbf s') p^{\mathbf a}(\mathbf s,\mathbf s')\Big)\Big]b^c_t(\mathbf s)\\
& \stackrel{(b)}{\le} \gamma \|Q^*\|_\infty,
\end{align*}
where $(a)$ follows from~(15)--(16) by using Cauchy-Schwartz inequality and the definition of sup-norm. $(b)$ holds due to the fact that  $\sum_{\mathbf s \in \ALPHABET S} \Big[ \sum_{\mathbf a \in \ALPHABET A} \pi^{\boldsymbol{\omega}}(\mathbf a| \mathbf s)\Big( \sum_{\mathbf s' \in \ALPHABET S} b^c_{t+1}(\mathbf s') p^{\mathbf a}(\mathbf s, \mathbf s')\Big)\Big]b^c_t(\mathbf s) \le 1$.
The last inequality implies contraction since $\gamma \in (0,1)$. This completes the proof of the lemma.
\end{proof}

Because  $\ALPHABET B^{\mu_t}$ and $\ALPHABET B^*$ are contractions,~\eqref{eq:Q-star-2} has a unique solution. Furthermore, since $r^{\mathbf a_t, \boldsymbol{\mathrm{br}}_t}$ is bounded, so is  $r^{\boldsymbol{\omega}_t}$ and consequently so is $Q^*$.

\subsection{Main result~1: Dynamic program}

The main result of this section is given by the following theorem, which provides a suitable dynamic program for the infinite horizon discounted reward \emph{dynamic team} problem and establishes the optimality of the joint-option policy. 
\begin{theorem}\label{thm:prescription-based-optimality}
For the $J$-agent Dec-POMDP  described above
\begin{enumerate}
  \item  The optimal state-value is the fixed point solution of the following dynamic program.
   \begin{equation}
V^*(b^c_t)
\DEFINED  \max_{\mu_t \in \ALPHABET M^+} \sum_{\boldsymbol{\omega}_t \in \Omega} \mu_t(\boldsymbol{\omega}_t|b^c_t) 
\Bigg[r^{\boldsymbol{\omega}_t}(b^c_t)
 + \gamma \sum_{\boldsymbol{\tilde o_t} \in \ALPHABET O \cup \{\varnothing\}} \PR(\boldsymbol{\tilde o_t}|b^c_t, \boldsymbol{\omega}_t) V^*(b^c_{t+1}) \Bigg], \label{eq:V-option}
\end{equation}
where $\ALPHABET M^+$ is the space of joint option-policies; $r^{\boldsymbol{\omega}_t}(b^c_t)$ is given by~\eqref{eq:r-omega}, $\PR(\boldsymbol{\tilde o_t}|b^c_t, \boldsymbol{\omega}_t)$, as given by~\eqref{eq:prob-o-tilde} is the observation-model and $b^c_{t+1}$ is given by~\eqref{eq:belief-update-t+1}.
\item 
Let $\ALPHABET M$ denote the space of Markov joint-option policies. Then, there exists a \emph{time-homogeneous}  Markov joint-option policy $\mu^* \in \ALPHABET M$ which is optimal, i.e.,
\[
\mu^* = \arg\max_{\mu_t \in \ALPHABET M} V^{\mu_t}(b^c_t),
\]
where $V^{\mu_t}$ is given by: 
  \begin{equation}
  V^{\mu_t}(b^c_t) = \sum_{\boldsymbol{\omega}_t \in \Omega} \mu_t(\boldsymbol{\omega}_t|b^c_t) \Bigg[r^{\boldsymbol{\omega}_t}(b^c_t)]
   + \gamma \sum_{\boldsymbol{\tilde o_t} \in \ALPHABET O \cup \{\varnothing\}} \PR(\boldsymbol{\tilde o_t}|b^c_t, \boldsymbol{\omega}_t)V^{\mu_t}(b^c_{t+1})\Bigg] \label{eq:V-mu}.
  \end{equation}
Then, 
$
V^*(b^c_t) = V^{\mu^*}(b^c_t).
$ and furthermore,  $\mu^*$ is obtained using the common belief $b^c_t$.
\end{enumerate}
\end{theorem}
\begin{proof}
\begin{enumerate}
\item 
As shown above, the system is a POMDP with $b^c_t$ acting as a state, so the state-value $V^{\mu_t}(b^c_t)$ for a given joint option-policy $\mu_t$ satisfies the Bellman equation given by~\eqref{eq:V-mu}.
It can be shown following standard results for
POMDP that~\eqref{eq:V-mu} is a contraction and hence there exists a unique bounded solution $V^{\mu_t}$.  

Since the set of probability measures on finite spaces is finite, we can use $\max$ instead of $\sup$ in defining the optimal state-value $V^*$ in~\eqref{eq:V-option}. Thus, we have  
\[
V^*(b^c_t) \DEFINED \max_{\mu_t \in \ALPHABET M^+} V^{\mu_t}(b^c_t).
\]
Since the maximum of a bounded function over a finite set is bounded, $V^*$ is unique and bounded.
\item Let $\mu^* \in \ALPHABET M$ be a time-homogeneous Markov joint option-policy. We need to show that such a $\mu^*$ exists. If it does, then  $V^* =  V^{\mu^*}$. The existence of a time-homogeneous Markov joint-option policy, which achieves the optimal state-value $V^*$, follows from \emph{Blackwell optimality}.\footnote{Blackwell optimality~\cite{blackwell1962} states that, in any MDP with finitely many states, finitely many actions and discounted returns, there is a pure stationary (time-homogeneous) strategy that is optimal, for every discount factor
close enough to one. An extension of Blackwell optimality holds for discounted infinite horizon POMDPs. See~\cite[Theorem~2.6.1]{krishnaArxiv2015} for details.}

Now, by~\eqref{eq:mu_t_psi_tilde} we can restrict our attention to the set of joint option-policies $M^{\psi, b^c_t}$ where any $\tilde \mu \in M^{\psi, b^c_t}$ is a function of the  coordination rule  $\psi$ and $b^c_t$. Thus, we have: 
\[
V^*(b^c_t) = \max_{\mu \in \ALPHABET M \cap \tilde M^{\psi, b^c_t}} V^\mu(b^c_t).
\]
which completes the proof. 
\end{enumerate}
\end{proof}
As a consequence of Theorem~\ref{thm:prescription-based-optimality}, we can now consider time-homogeneous Markov option policies $\mu$. Subsequently, we use  $\pi^{\boldsymbol{\omega}}$, $\pi^{b,\boldsymbol{\omega}}$ and $\beta^{\boldsymbol{\omega}}$ in the rest of the paper. 

Note that planning with a factored common belief reduces the exponential computation complexity to polynomial. Let the cardinality of a finite factored state space $\ALPHABET S = \ALPHABET S^1 \times \dots \times \ALPHABET S^J$ is $|\ALPHABET S| = \prod_{j \in \ALPHABET J}|\ALPHABET S^j|$. Similarly, let the cardinality of a finite factored action space $\ALPHABET A = \ALPHABET A^1 \times \dots \times \ALPHABET A^J$ is $|\ALPHABET A| = \prod_{j \in \ALPHABET J}|\ALPHABET A^j|$. Then, at each iteration of policy iteration the computational complexity is $O(|\ALPHABET S|^2 |\ALPHABET A|(|\ALPHABET S| + |\ALPHABET A|)$, which is exponential in the number of agents $J$. In contrast, with factored agents and belief, the computational complexity becomes $O(|\ALPHABET S^j|^2 |\ALPHABET A^j| ((J-1)J |\ALPHABET S^j| + J|\ALPHABET A^j|))$ for fixed $j$, which is polynomial in $J$ and thus scalable.

\section{Learning in Dec-POMDPs with options}\label{sec:learning}

We are interested in individual agents learning independent policies and so we concentrate on learning the best factored actor for a domain, even if it is suboptimal in a global sense. Also, for ease of readability, in this section we use full observability in the derivations, i.e., $\pi^{b,\omega^j_t}(\mathrm{br}^j_t|o^j_t) = \pi^{b,\omega^j_t}(\mathrm{br}^j_t|s^j_t)$ and $\pi^{\omega^j_t}(a^j_t|o^j_t) = \pi^{\omega^j_t}(a^j_t|s^j_t)$. 
However, our results hold even if the agents are not locally fully observable, where the observations depend on the states probabilistically (as is discussed in the previous sections) or deterministically (e.g., state-embedding as we use in our experiments).

Our proposed algorithm for learning options,  \emph{Distributed Option Critic} (DOC; see Algo.\ref{alg:DOC}), 
employs the \emph{option-critic} architecture~\cite{Bacon2017TheOA},  leverages the assumption of \emph{factored actions} of agents in the distributed policy updates and utilizes the \emph{sufficient statistic of common information} to learn the critic. The centralized option evaluation is presented from the coordinator's point of view, where the centralized \emph{option-value} $Q$ and \emph{intra-option value} $Q_{\text{intra}}$ using a joint-state (embedding) of all agents inferred from the \emph{common belief} and the common information of options and actions of all agents. The agents learn to complete a cooperative task by learning in a model-free way. In the \emph{centralized option evaluation} step, the centralized critic (coordinator) evaluates in \emph{temporal difference} manner~\cite{Apostol_TD} the performance of all agents via a shared reward (plus a broadcast penalty in case of costly communication) using the common information. Once any agent~$j$ terminates its own option $\omega^j$, it chooses a new option in a greedy manner using a \emph{modified} critic value by replacing its own component of the current joint option $\boldsymbol{\omega}$ with a new available option and keeping all other components $\boldsymbol{\omega}^{-j}$ unchanged. 

Each agent updates its parameterized intra-option policy, broadcast policy and termination function through \emph{distributed option improvement}. In order to learn their policies, each agent~$j$ uses a \emph{modified critic value}, obtained by replacing $j$-th  component of the common information (the joint-embedding $\mathbf s_k$ sampled from the common belief $b^c_k$) in the critic $Q$ and intra-option value $Q_{\text{intra}}$ with their private information (their own embedding $s^j$) and keeping all other components $\mathbf s^{-j}$ intact\footnote{Algorithms for COE and DOI are in the appendix.}.

\begin{algorithm2e}[!tb]
\def\1#1{\hbox to 1em{\hfill$\mathsurround0pt #1$}}
\SetKwInOut{Input}{Input}
\SetKwInOut{Output}{Output}
\SetKwInOut{Init}{initialize}
\SetKwFor{ForAll}{forall}{do}{}
\SetKwRepeat{Do}{do}{while}
\DontPrintSemicolon
\Input{Set of goals $\ALPHABET G$; broadcast penalty $B$ (for intermittent broadcast); learning rates $\alpha_{\theta^j}$, $\alpha_{\epsilon^j}$, $\alpha_{\phi^j}$ and $\alpha_Q$; pool of options $\Omega$; number of episodes $\mathsf{N}_{\text{epi}}$;}
\Output{Estimate $Q$ of the optimal option-value $Q^*$}
\For{episode in $\mathsf{N}_{\text{epi}}$}{
\Init{pool of available options $\Omega_{\text{avail}}= \Omega$;
initial common belief $b_0$ (or initial common information $\ALPHABET I^c_0$); parameters $\theta^j$, $\epsilon^j$ and $\phi^j$, $j \in \ALPHABET J$\;
}
\For{iteration $k = 1$ upto end of episode}{
  \label{step:mu} \textbf{Choose joint-option} $\boldsymbol{\omega}$ based on softmax or epsilon-greedy option-policy $\mu$. 
  Denote the true current joint-state by $\mathbf s$. Choose action $\mathbf a_k =(a^1_k, \dots, a^J_k)$ in true current joint-state $\mathbf s$; $a^j_k \sim \pi^{\omega^j_k, \theta^j}$. \textbf{Take a step through environment} and get a reward $r$.\;
  \textbf{Sample broadcast action} $\mathrm{br}^j$, $j \in \ALPHABET J$ (for intermittent broadcast; otherwise $\mathrm{br}^j = 1$).\;
  \textbf{Get a new joint-observation} $\boldsymbol{\tilde o}_k$.\;
Do a \emph{centralized option-value evaluation} to \textbf{compute} $\mathbf{Q}$. If an agent's option terminates, choose a new option greedily keeping all other agents' options frozen. \;
\textbf{Update action-policy, broadcast-policy and termination parameters} $(\theta^j, \epsilon^j, \phi^j)$ using \emph{distributed option improvement}. To do so, replace component $s^j_k$ of $\mathbf s_k \sim b^c_k$ by $s^j$, keep all other components $\mathbf s_k^{-j}$ frozen and plug into $Q$.\;
}
}
\Return{$Q$}

\caption{Distributed Option Critic (DOC)}\label{alg:DOC}
\end{algorithm2e}

\begin{figure*}[t]
        \centering
        \begin{subfigure}[b]{0.26\textwidth}
     \includegraphics[width=0.7\linewidth]{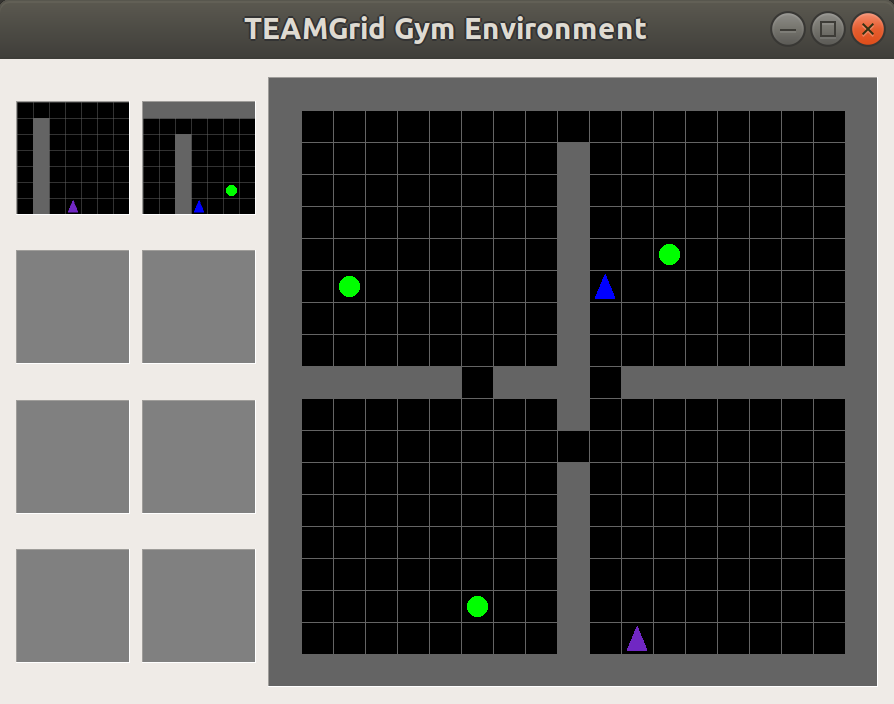}
                \caption{}
                \label{fig:eps03}
        \end{subfigure}
        \begin{subfigure}[b]{0.28\textwidth}  
        \includegraphics[width=0.8\linewidth]{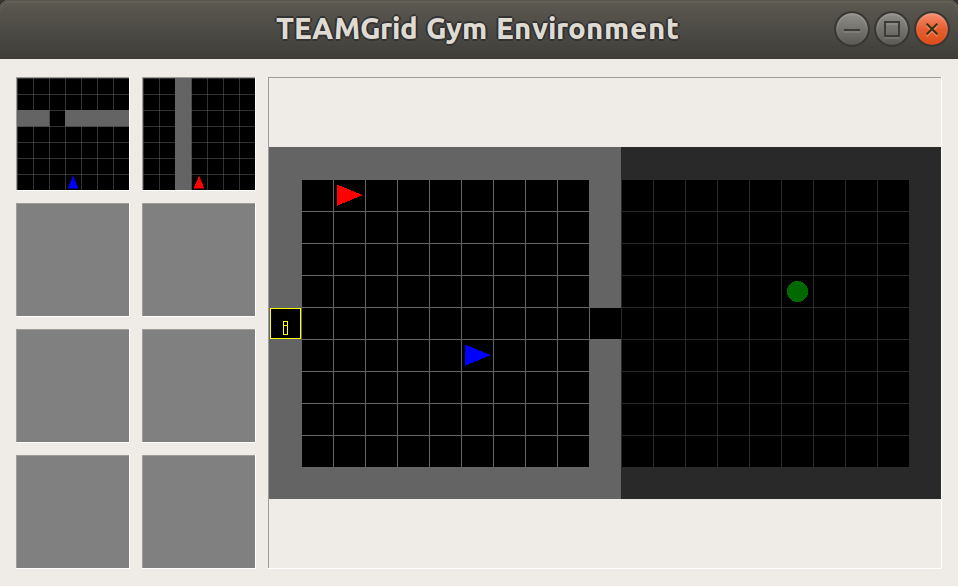}
                \caption{}
              \label{fig:eps0}
        \end{subfigure}%
        ~ 
        \begin{subfigure}[b]{0.28\textwidth}
     \includegraphics[width=0.8\linewidth]{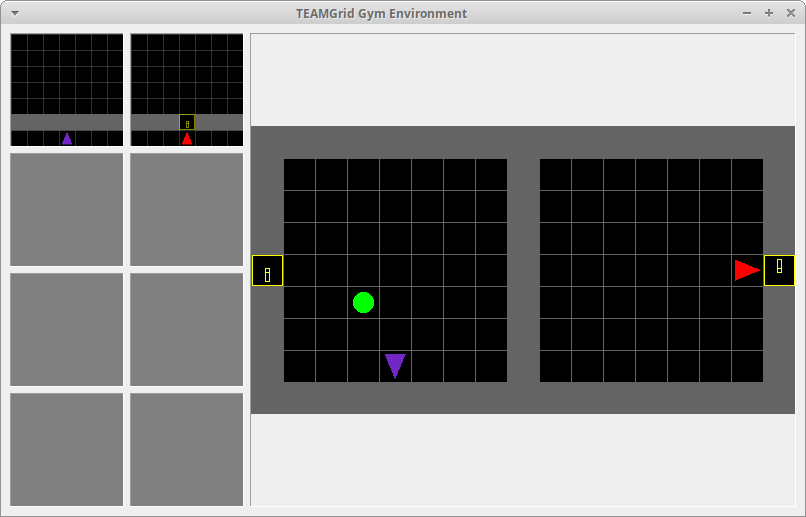}
                \caption{}
                \label{fig:eps03}
        \end{subfigure}
        \caption{\emph{TEAMGrid} environments: (a) FourRooms, (b) Switch and (c) DualSwitch.}\label{fig:teamgrid_envs}
\end{figure*}
The action-policy, broadcast-policy and the termination function of Agent~$j$ are parameterized by $\theta^j$, $\epsilon^j$ and $\varphi^j$ respectively and are learnt in distributed manner in the Distributed Option Improvement step of DOC, through stochastic gradient descent.

\subsection{Main result~2: Convergence of DOC}
Using arguments for the convergence of the policy-gradient based algorithms (e.g.,~\cite{SuttonPolicyGradient:1999}) and the local optima achieved by distributed stochastic gradient descent~\cite[Theorem~1]{PeshkinEtAl}, we can show that
DOC converges to the optimal option-value $Q^*$.
The proof relies on first arguing that for factored agents, the distributed stochastic gradient leads to local optima in the dynamic cooperative game, and then showing that the expected value of the option-value update in DOC is a contraction, leading to convergence to the optimal option-value. 
We first state the following lemma.
\begin{lemma}\label{lemma:DOC-SPNE}
Distributed gradient descent in a cooperative Dec-POMDP with options and with factored agents leads to local optima. 
\end{lemma}
\textbf{Proof sketch:} According to~\cite[Theorem~1]{PeshkinEtAl}, for factored agents, distributed gradient descent is equivalent to joint gradient descent and thus achieves local optima. Then the lemma follows by~\cite[Theorem~1]{PeshkinEtAl} due to the fact that DOC is a distributed gradient descent and so it leads to local optima. 
\begin{figure*}[t]
        \centering
        \begin{subfigure}[b]{0.3\textwidth}
           \includegraphics[width=0.95\linewidth]{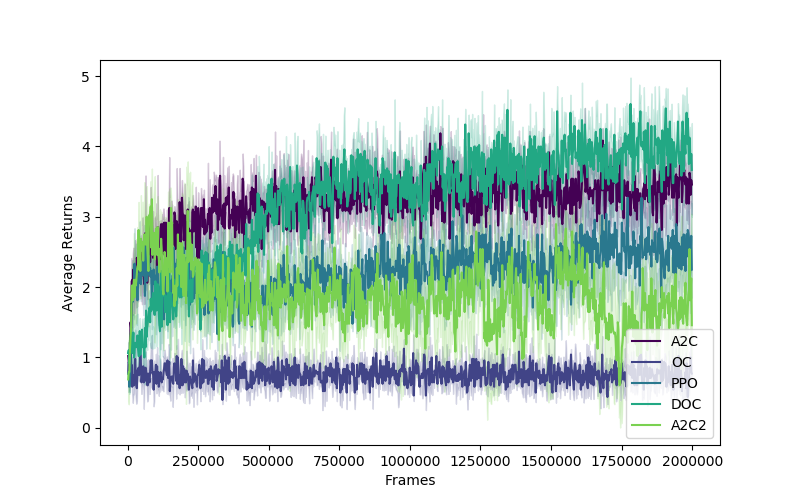}
                \caption{}
              \label{fig:2a_3g}
        \end{subfigure}%
        \begin{subfigure}[b]{0.3\textwidth}
             \includegraphics[width=0.95\linewidth]{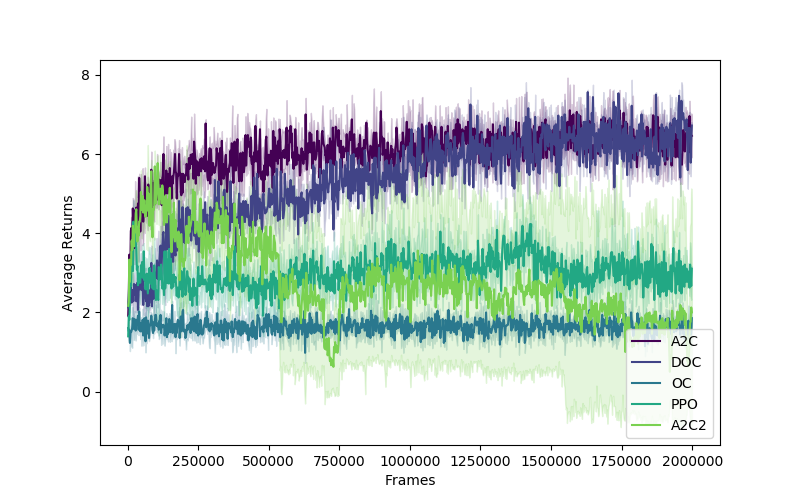}
                \caption{}
                \label{fig:3a_5g}
        \end{subfigure}%
        \begin{subfigure}[b]{0.275\textwidth}
              \includegraphics[width=0.95\linewidth]{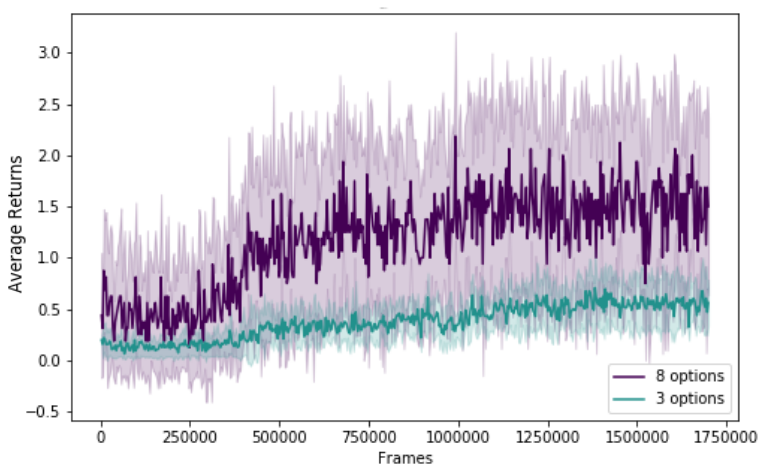}
                \caption{}
             \label{fig:num_options}
        \end{subfigure}
        \\
        \begin{subfigure}[b]{0.3\textwidth}
              \includegraphics[width=0.95\linewidth]{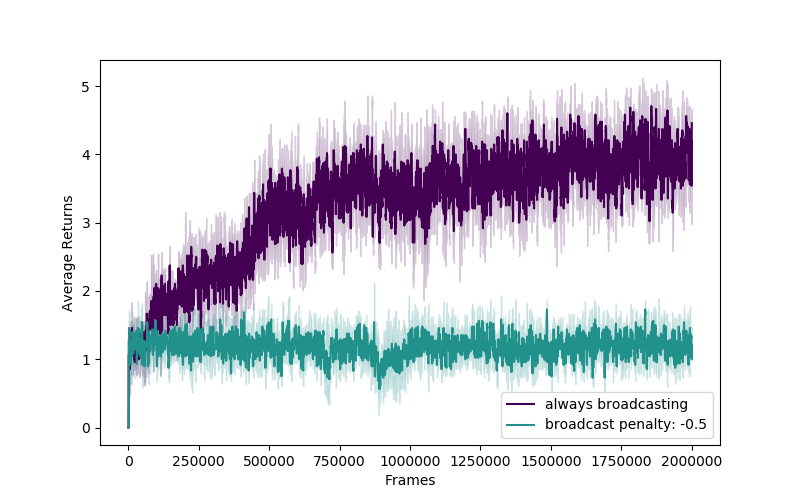}
                \caption{}
             \label{fig:bp_comp}
        \end{subfigure}
        \begin{subfigure}[b]{0.27\textwidth}
              \includegraphics[width=0.95\linewidth]{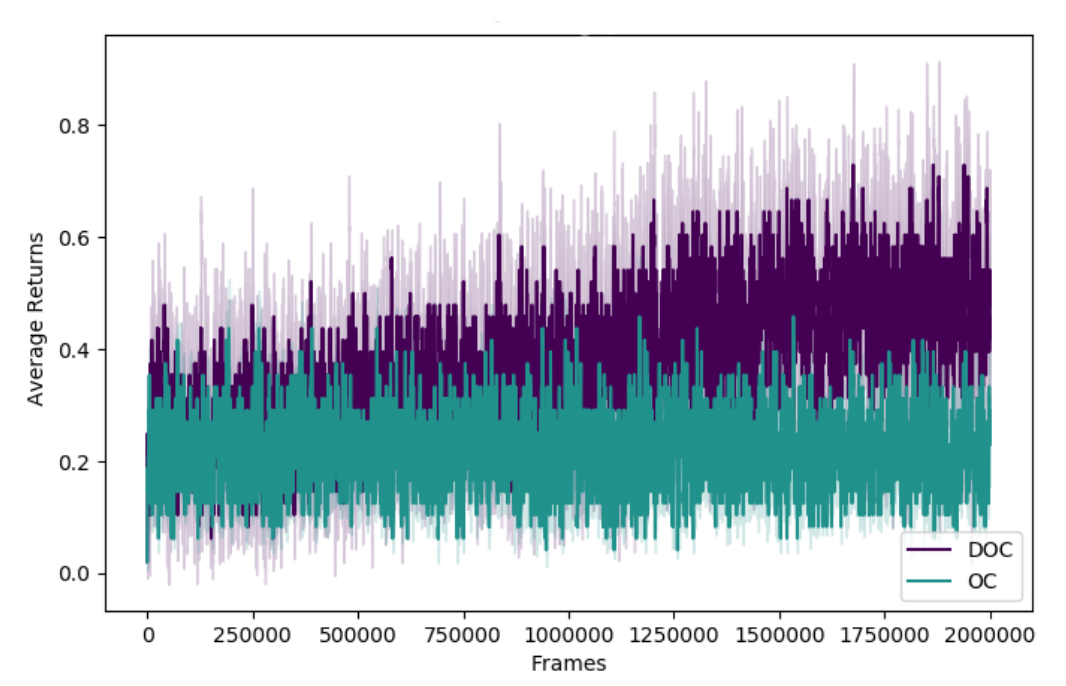}
                \caption{}
             \label{fig:switch}
        \end{subfigure}
        \begin{subfigure}[b]{0.3\textwidth}
              \includegraphics[width=0.95\linewidth]{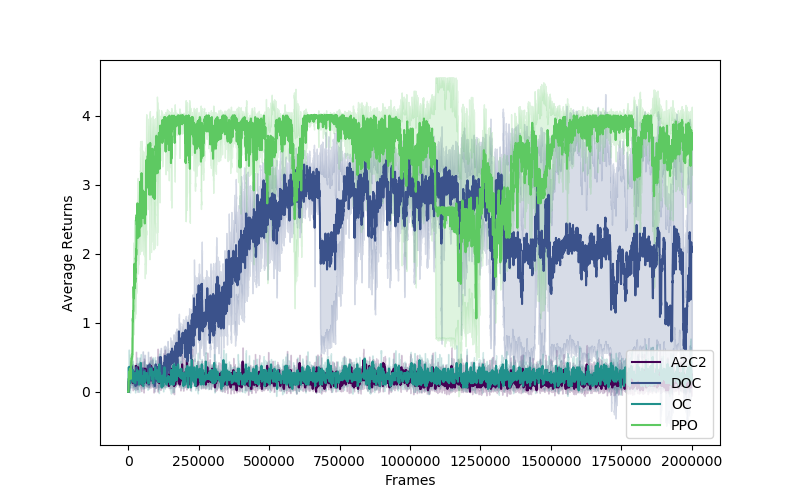}
                \caption{}
             \label{fig:dual_switch}
        \end{subfigure}
        \caption{\emph{TEAMGrid} results. (a) FourRooms average returns with 2 agents and 3 goals, (b) FourRooms average returns with 3 agents and 5 goals (c) FourRooms DOC: increasing number of options improved average returns, (d) FourRooms DOC average returns with always broadcasting (broadcast penalty 0.0) and intermittent broadcasting (broadcast penalty = -0.5), (e) Switch average returns, (f) DualSwitch average returns.}\label{fig:4room_average_returns}
\end{figure*}

\begin{figure}
\centering
\includegraphics[width=0.5\linewidth]{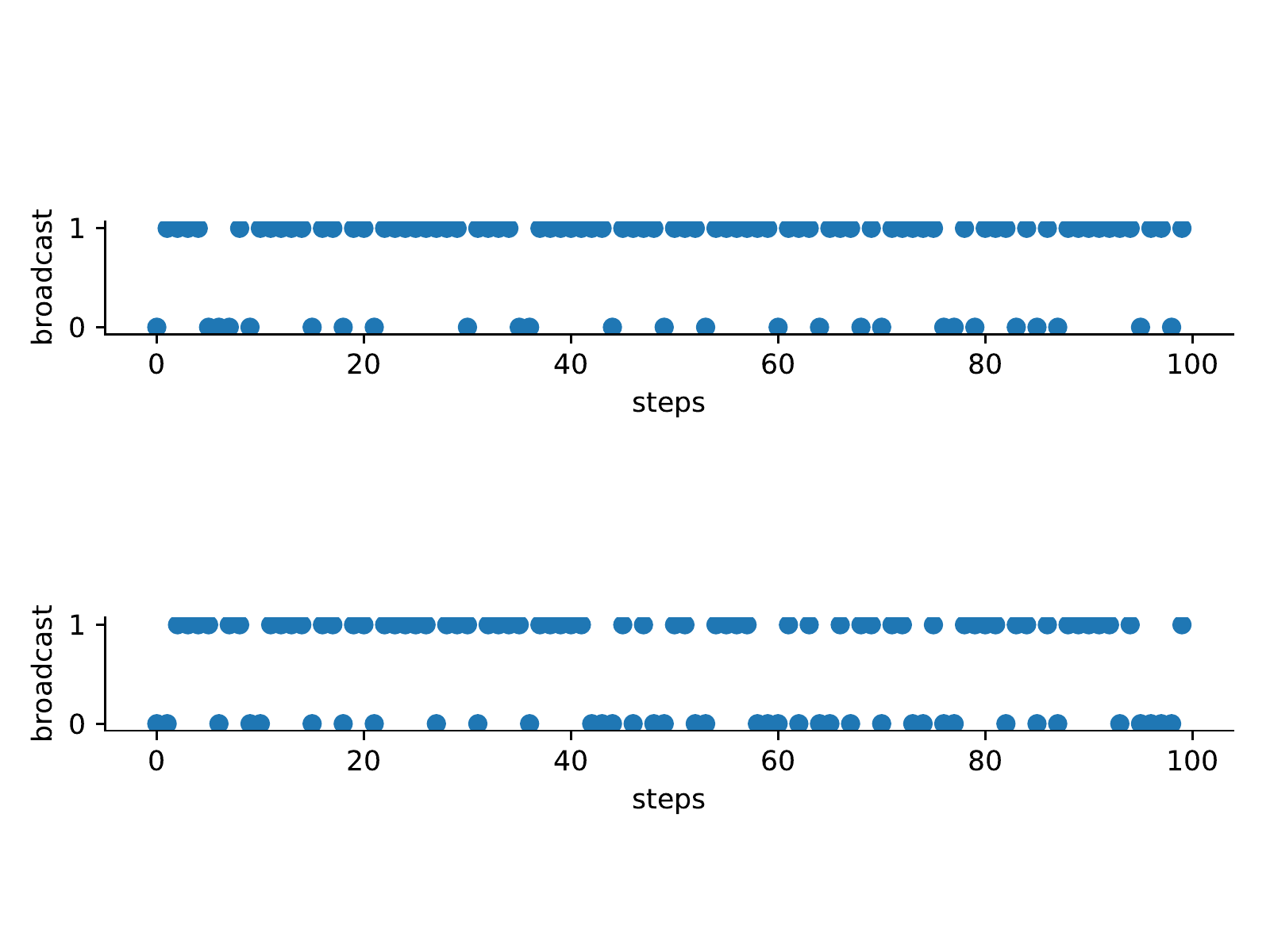}
 \caption{Intermittent broadcast: increase in broadcast penalty reduces the frequency of broadcast.}
  \label{fig:brd_freq}
\end{figure}

\begin{proof}[Convergence of DOC]
We now show that for the learning problem, intra-option $Q$-learning using common belief converges almost surely to the optimal $Q$-values, $Q^*$, for every joint-option $\boldsymbol{\omega}_k \in \Omega$, regardless of what options are executed during learning, provided that every action gets executed in every state infinitely often. For every joint-option $\boldsymbol{\omega}_k$, a joint-action $\mathbf a_k$ and broadcast $\boldsymbol{\mathrm{br}}_k$ is chosen according to action-policy $\pi^{\boldsymbol{\omega_k}}$ and broadcast policy $\pi^{b, \boldsymbol{\omega_k}}$ respectively and then an off-policy one-step TD update is executed as follows.
\[
  Q(\mathbf s_k, \boldsymbol{\omega}_k) = Q(\mathbf s_k, \boldsymbol{\omega}_k) + \alpha_Q \delta, 
\] 
where $\delta$ is the TD-error given by
\[
  \delta = r^{\boldsymbol{\omega}_k}(\mathbf s) + \gamma U(\mathbf s_{k+1},\boldsymbol{\omega}_k) - Q(\mathbf s_k,\boldsymbol{\omega}_k),
\]
where $\mathbf s$ is the true joint-state. At each step $k$, the joint-states $\mathbf s_k$ and $\mathbf s'_{k+1}$ are sampled from the common beliefs $b^c_k$ and $b^c_{k+1}$ respectively.
First we show that the expected value of $\delta$ equals $r^{\boldsymbol{\omega}_t}(b^c_k) + \gamma \EXP[U(b^c_{k+1}, \boldsymbol{\omega}_t)\,|\, b^c_k] - Q(b^c_k, \boldsymbol{\omega}_k)$. Note that by definition as given by~\eqref{eq:belief-update-t+1}, $b^c_{k+1}$ gives the belief of the true next joint-state $\mathbf s'$. Then, we have
\begin{align*}
 \EXP[\delta |b^c_k] 
  &= \sum_{\mathbf s \in \ALPHABET S}
  \sum_{\boldsymbol{\mathrm{br}}_k \in \{0,1\}^J} 
\sum_{\mathbf a_k \in \ALPHABET A} 
\pi^{b,\boldsymbol{\omega}_t}(\boldsymbol{\mathrm{br}}_k|\mathbf s)\pi^{\boldsymbol{\omega}_k}(\mathbf a_k|\mathbf s) r^{\mathbf a_k, \boldsymbol{\mathrm{br}}_k}(\mathbf s)b^c_k(\mathbf s) \\
& \hskip 2em + \sum_{\mathbf s_k \in \ALPHABET S} \sum_{\boldsymbol{\mathrm{br}}_k \in \{0,1\}^J} 
\sum_{\mathbf a_k \in \ALPHABET A} 
\pi^{b,\boldsymbol{\omega}_t}(\boldsymbol{\mathrm{br}}_k|\mathbf s_k)\pi^{\boldsymbol{\omega}_k}(\mathbf a_k|\mathbf s_k)
 \Bigg[\gamma 
  \sum_{\mathbf s' \in \ALPHABET S} 
  p^{\mathbf a_k}(\mathbf s_k,\mathbf s') U(\mathbf s', \boldsymbol{\omega}_k)
 - Q(\mathbf s_k, \boldsymbol{\omega}_k)\Bigg]b^c_k(\mathbf s_k)\\
 &\stackrel{(a)}{=} r^{\boldsymbol{\omega}_k}(b^c_k) + \gamma \EXP[U(b^c_{k+1}, \boldsymbol{\omega}_k)\,|\,b^c_k] - Q(b^c_k,\boldsymbol{\omega}_k),
\end{align*}
where $(a)$ holds by the definitions of $r^{\boldsymbol{\omega}_k}(\mathbf s)$, $b^c_{k+1}$, $U(b^c_{k+1},\boldsymbol{\omega}_k)$ and  $Q(b^c_k,\boldsymbol{\omega}_k)$.

Next, note that the by definition of intra-option $Q$-learning with full observability (e.g. see \cite[Theorem~3]{SuttonEtAl1999}), we have that for any $\varepsilon \in \reals_{>0}$,
\begin{equation}\label{eq:Q-convergence}
  \max_{\mathbf s'', \boldsymbol{\omega}''} |Q(\mathbf s'', \boldsymbol{\omega}'') - Q^*(\mathbf s'', \boldsymbol{\omega}'')| < \varepsilon.
\end{equation}
The rest of the proof follows by showing that the expected value of $r^{\boldsymbol{\omega}_k}(\mathbf s) + \gamma U(\mathbf s'_{k+1}, \boldsymbol{\omega}_k)$ converges to $Q^*$, which is given as follows. 
For ease of exposition, we drop the subscript $k$ everywhere except for common beliefs in the following derivation.
\begin{align*}
&\bigm|r^{\boldsymbol{\omega}}(b^c_k) + \gamma \EXP[U(b^c_{k+1}, \boldsymbol{\omega})\,|\, b^c_k] - Q^*(b^c_k,\boldsymbol{\omega})\bigm| \\
& = \Biggm|\gamma
\sum_{\boldsymbol{\mathrm{br}} \in \{0,1\}^J}\sum_{\mathbf a \in \ALPHABET A} \sum_{\mathbf s \in \ALPHABET S}
\pi^{b,\boldsymbol{\omega}}(\boldsymbol{\mathrm{br}}|\mathbf s)\pi^{\boldsymbol{\omega}}(\mathbf a|\mathbf s) 
\bigg(\sum_{\mathbf s' \in \ALPHABET S} p^{\mathbf a}(\mathbf s, \mathbf s')
\bigg[ \beta^{\mathbf o}_{\text{none}}(\mathbf s')\Big(Q(\mathbf s', \boldsymbol{\omega}) - Q^*(\mathbf s',\boldsymbol{\omega})\Big) \\
& \hskip 2em + \big(1-\beta^{\boldsymbol{\omega}}_{\text{none}}(\mathbf s')\big) \Big(\max_{\ALPHABET T \in \text{Pow}(\ALPHABET J)}\max_{\boldsymbol{\omega}' \in \Omega_{\text{avail}}(\ALPHABET T)} Q(\mathbf s',\boldsymbol{\omega}') 
- \max_{\ALPHABET T \in \text{Pow}(\ALPHABET J)}\max_{\boldsymbol{\omega}' \in \Omega_{\text{avail}}(\ALPHABET T)} Q^*(\mathbf s',\boldsymbol{\omega}') \Big) \bigg] \bigg) b^c_t(\mathbf s)\Biggm|\\
&\stackrel{(a)}{\le} \gamma \sum_{\mathbf s \in \ALPHABET S}
\sum_{\boldsymbol{\mathrm{br}} \in \{0,1\}^J}\sum_{\mathbf a \in \ALPHABET A} 
\pi^{b,\boldsymbol{\omega}}(\boldsymbol{\mathrm{br}}|\mathbf s)\pi^{\boldsymbol{\omega}}(\mathbf a|\mathbf s) 
\bigg( \sum_{\mathbf s' \in \ALPHABET S} p^{\mathbf a}(\mathbf s,\mathbf s')
 \bigg[\max_{\mathbf s'', \boldsymbol{\omega}''} \bigm| Q(\mathbf s'',\boldsymbol{\omega}'') - Q^*(\mathbf s'',\boldsymbol{\omega}'')\bigm| \bigg]\bigg) b^c_t(\mathbf s)\\
& \stackrel{(b)}{<} \varepsilon \gamma \sum_{\mathbf s \in \ALPHABET S}
\sum_{\boldsymbol{\mathrm{br}} \in \{0,1\}^J}\sum_{\mathbf a \in \ALPHABET A} 
\pi^{b,\boldsymbol{\omega}}(\boldsymbol{\mathrm{br}}|\mathbf s)\pi^{\boldsymbol{\omega}}(\mathbf a|\mathbf s) \Big(\sum_{\mathbf s' \in \ALPHABET S} p^{\mathbf a}(\mathbf s,
\mathbf s')\Big)b^c_t(\mathbf s) \\
& \stackrel{(c)}{\le} \varepsilon \gamma.
\end{align*}
Note that since $\ALPHABET J$ is finite, so is $\text{Pow}(\ALPHABET J)$. Consequently, $(a)$ holds since maximum over a finite set is bounded and since maximum over real line is convex. $(b)$ holds by~\eqref{eq:Q-convergence} and $(c)$ holds since for fixed $\mathbf a$ and $\mathbf s$, $\sum_{\mathbf s' \in \ALPHABET S} b^c_{t+1}(\mathbf s') p^{\mathbf a}(\mathbf s,\mathbf s') \le 1$; for a fixed $\mathbf s$, $\sum_{\boldsymbol{\mathrm{br}}\in \{0,1\}^J}\sum_{\mathbf a \in \ALPHABET A}\\ \pi^{b,\boldsymbol{\omega}}(\boldsymbol{\mathrm{br}}|\mathbf s)\pi^{\boldsymbol{\omega}}(\mathbf a|\mathbf s) \le 1$ and $\sum_{\mathbf s \in \ALPHABET S} b^c_t(\mathbf s) = 1$. The last inequality implies convergence since $\varepsilon$ can be arbitrarily small and $\gamma \in (0,1)$.

The convergence of intra-option $Q$-learning in teams along with Lemma~\ref{lemma:DOC-SPNE} ensures that the option-value $Q$ obtained by DOC  converges to the optimal option-value $Q^*$.  
\end{proof}

\section{Experiments}\label{sec:experiments}


We empirically evaluate the merits of DOC in cooperative multi-agent tasks, and compare it to its single-agent counterpart, Option-Critic (OC), Advantage Actor-Critic (A2C) and Proximal Policy Optimization (PPO). We also extend A2C to the multi-agent setting by incorporating a centralized critic that uses common information. To the best of our knowledge, this is equivalent to Counterfactual Multi-Agent Policy Gradients (COMA) when the individual agents broadcast at every time step. We call this algorithm A2C2.

In all of our experiments, we use deep neural networks for actors and critics. These networks use  two linear layers with 64 hidden units and \emph{tanh} activation. The memory for both the actor and the critic are implemented with \emph{Long Short-Term Memory}
(LSTM) cells~\cite{Hochreiter_lstm} as they allow a natural way to incorporate observations into a latent state. 
All neural networks were optimized using \emph{RMSProp}\cite{rmsprop} and \emph{Adam}\cite{kingma_ba15}, which use adaptive learning rates for stochastic gradient descent. 
Experiments were run using CPU cores and the mean and variances were computed using 3 to 5 seeds over 40000 episodes (equiv. to 2-million frames) on \emph{TEAMGrid}. 
 
 \subsection{TEAMGrid: A multi-agent extension of Minigrid}
 We created \emph{TEAMGrid} environments that extend Minigrid \cite{gym_minigrid} to incorporate the multi-agent setting. An  illustration of our environment is given in (Fig.~\ref{fig:teamgrid_envs}). In each environment (a,b and c), the bigger frame to the right displays the global perspective of the environment where we see agents (triangles), goals (circles) and switches (yellow squares on the walls). The smaller frames to the left of each environment displays the local perspective of each agent's field of view. 
 Similar to Minigrid~\cite{gym_minigrid}, the states of the agents are their positions, their observations are the cells within their fields of view and the available actions are \emph{Left, Right, Forward, Toggle, Pickup, Drop}. 
 The agents collect sparse rewards upon completing the task (e.g., discovering goals, picking up a ball, toggling a switch). 
 
In \textbf{\emph{TEAMGrid-FourRooms}} (Fig.~\ref{fig:teamgrid_envs}a), several agents try to find one or more goals while avoiding collision with each other (collision incurs a penalty). In \textbf{\emph{TEAMGrid-Switch}} (Fig.~\ref{fig:teamgrid_envs}b), two agents are placed in two rooms. There is a goal object in the room on the right. The room on the right is dark until the switch in the room on the left is turned on. To maximize efficiency, one agent should go in the room on the right while the other turns on the switch in the room on the left. In \textbf{\emph{TEAMGrid-DualSwitch}} (Fig.~\ref{fig:teamgrid_envs}c), two agents are placed in two rooms, each with a switch and a goal. When one agent turns on a switch, the goal in the other room appears. The task is to get all goals. 

\subsection{Results}
 Fig.~\ref{fig:4room_average_returns}a,b show that DOC outperforms the baselines in \emph{FourRooms}, demonstrating the benefit of options in this environment. We also notice that DOC's performance isn't affected when we scale up the number of agents and goals. This is in contrast with A2C2's performance.  In \emph{Switch} (Fig.~\ref{fig:4room_average_returns}e) DOC outperforms OC. In fact, OC never shows an increase in reward since this environment requires cooperation whereas DOC manages to can capture this.
 In \emph{DualSwitch} (Fig.~\ref{fig:4room_average_returns}f), PPO outperforms DOC. We suspect that since the agents act simultaneously, it was not particularly beneficial for temporal abstraction, the merit of which is mostly reflected in sequential action execution. However, PPO's performance fluctuates quite significantly (we believe this is because the agents explore independently to get to the goal without any communication among themselves), while DOC performs competitively and its performance is comparatively more consistent. Both PPO and DOC do significantly better than A2C2 and OC.

We also ran experiments to investigate the effect that applying a broadcast penalty will have on both the frequency of broadcasts and performance. We find that the agents' performance were heavily attenuated by intermittent broadcasting compared to always broadcasting (Fig.~\ref{fig:4room_average_returns}d). Communication in distributed networks is a challenging problem mainly due to the losses in the channels and there exits a fundamental trade-off between communication cost and the estimation accuracy. The optimality of distributed communication depends to a great extent on generating a reliable estimate of the information states of the agents. Our agents learn to broadcast using estimates of others' embeddings based on the common information and using the  broadcast penalty as the feedback signal. In \emph{FourRooms}, they were communicating 61\% of times as opposed to 74\% of times when the broadcast penalty changed from -0.01 to -0.5, as is shown in Fig.~\ref{fig:brd_freq}.

Finally, we study the effect of changing the number of options for DOC (Fig.~\ref{fig:4room_average_returns}c). We see that increasing in the number of options improves the overall performance. Interestingly, in practice we notice that changes to the number of options should be met with proportional changes to the amount exploration (i.e. entropy regularization) to see these improvements in performance. We believe this is due to the fact that having more options allows agents to increase the amount of targeted learning however to learn these abstract targets, more exploration in necessary. 



\section{Conclusion}\label{sec:discussion}
In this paper, we extend the options framework for temporal abstraction to Dec-POMDPs for cooperative multi-agent systems. We leverage the common information approach in tandem with temporal abstraction and use it to convert the Dec-POMDP to an equivalent POMDP. We then show that the corresponding planning problem has a unique solution. 
We also propose DOC, a model free algorithm for learning options. We show that DOC leads to local optima and analyze its asymptotic convergence. The implication of Lemma~\ref{lemma:DOC-SPNE} and the convergence of DOC is that DOC results in local optima $\boldsymbol{\omega}^* \DEFINED (\omega^{1*}, \dots, \omega^{J*})$, where $\omega^{j*}$ is achieved by $\pi^{j*}$, $\pi^{b,j*}$ and $\beta^{j*}$. 
We create a platform for gridworld environments facilitating multi-agent framework. Finally, our empirical results show that DOC performs competitively against the baselines. 

As a future work, we would like to compare our method with the contemporary research on multi-agent temporal abstraction, some of which we have mentioned in the introduction. Also, we aim to test the performance of DOC in other environments suitable for multi-agent setting. Lastly, communication in a distributed environment is hard due to unreliability of the communication channels (e.g., packet drops in the channels) and so learning to communicate optimally is a non-trivial problem by itself. In our work the agents learn to broadcast to all other agents using a broadcast penalty. Learning to communicate only to \emph{neighbors}, learning some characteristics of the channel (e.g. probability of packet drops) and communication with partial knowledge of the channel (e.g. with some side information about the channel) are interesting areas of future research.

\appendix
\section{Appendix: Centralized option evaluation and distributed option improvement}
Algo.~\ref{alg:COE} describes centralized option evaluation and Algo.~\ref{alg:DOI} describes distributed option improvement using policy gradient method.

\begin{algorithm2e}[!tb]
\def\1#1{\hbox to 1em{\hfill$\mathsurround0pt #1$}}
\SetKwInOut{Input}{Input}
\SetKwInOut{Output}{Output}
\SetKwInOut{Init}{initialize}
\SetKwProg{Fn}{function}{}{}
\SetKwFor{ForAll}{forall}{do}{}
\SetKwRepeat{Do}{do}{while}
\DontPrintSemicolon
\Fn{OE($r$, $\mathbf{s}_k$, $\boldsymbol{\omega}$, $\mathbf{a}_k$)}{
\Input{per-step reward $r$;  joint-state $\mathbf{s}_k$, joint-option $\boldsymbol{\omega}$ and joint-action $\mathbf{a}_k$}
\Output{$Q_{\text{intra}}(\mathbf{s}_k, \boldsymbol{\omega}, \mathbf{a}_k)$, $Q(\mathbf s_k, \boldsymbol{\omega})$}
Compute the TD-error $\delta$ as follows:
    \[
      \delta = r. 
    \]
    If $\mathbf s'$ is not terminal, do:
        \[
        \delta \leftarrow \delta + U(\mathbf s'_k,\boldsymbol{\omega}),
        \]
    where 
     \begin{align*}
  U(\mathbf s'_k, \boldsymbol{\omega})  &\DEFINED \beta^{\boldsymbol{\omega}}_{\text{none}}(\mathbf s')  Q(\mathbf s'_k, \boldsymbol{\omega}) \\
   & + (1-\beta^{\boldsymbol{\omega}}_{\text{none}}(\mathbf s')) \max_{\ALPHABET T \in \mathrm{Pow}(\ALPHABET J)} \max_{\boldsymbol{\omega}' \in \Omega(\ALPHABET T)} Q(\mathbf s'_k, \boldsymbol{\omega}'),
  \end{align*}
  where $\Omega(\ALPHABET T)$ denotes the set of options for agents in $\ALPHABET T \subseteq \ALPHABET J$. \\
  $Q(\mathbf s'_k, \boldsymbol{\omega}' )$ is given by:
  \[
  Q(\mathbf s'_k, \boldsymbol{\omega}' ) \DEFINED
  \sum_{\mathbf a_k \in \ALPHABET A} \pi^{\boldsymbol{\omega}'}(\mathbf a_k| \mathbf s'_k) Q_{\text{intra}}(\mathbf s'_k, \boldsymbol{\omega}', \mathbf a_k),
  \]
  and $Q_{\text{intra}}(\mathbf s'_k, \boldsymbol{\omega}', \mathbf a_k)$ is given by:
  \[
  Q_{\text{intra}}(\mathbf s'_k, \boldsymbol{\omega}', \mathbf a_k) \DEFINED
  r^{\mathbf a_k}(\mathbf s'_k)
  + \gamma \sum_{\mathbf s'' \in \ALPHABET S} p^{\mathbf a_k}(\mathbf s'_k, \mathbf s'')U(\mathbf s'', \boldsymbol{\omega}')
  \]\;
  Update $\delta$ as follows:
  \[
  \delta \leftarrow \delta - Q_{\text{intra}}(\mathbf{s}_k, \boldsymbol{\omega}, \mathbf{a}_k)
  \]\;
  Update $Q_{\text{intra}}(\mathbf{s}_k, \boldsymbol{\omega}, \mathbf a_k)$ as follows:
  \[
   Q_{\text{intra}}(\mathbf s_k, \boldsymbol{\omega}, \mathbf a_k) \leftarrow Q_{\text{intra}}(\mathbf s_k, \boldsymbol{\omega}, \mathbf a_k) + \alpha_Q \delta
  \]\;
\Return{$Q_{\text{intra}}(\mathbf{s}_k, \boldsymbol{\omega}, \mathbf{a}_k)$, $Q(\mathbf s_k, \boldsymbol{\omega})$}
}
\caption{Centralized Option Evaluation (COE)}\label{alg:COE}
\end{algorithm2e}

\begin{algorithm2e}[!tb]
\def\1#1{\hbox to 1em{\hfill$\mathsurround0pt #1$}}
\SetKwInOut{Input}{Input}
\SetKwInOut{Output}{Output}
\SetKwInOut{Init}{initialize}
\SetKwProg{Fn}{function}{}{}
\SetKwFor{ForAll}{forall}{do}{}
\SetKwRepeat{Do}{do}{while}
\DontPrintSemicolon
\Fn{DOI($\mathbf{s}_k$, $\mathbf o$, $\mathbf{a}_k$, $Q_{\text{intra}}$, $Q$)}{
\Input{$r$, $\mathbf{s}_k$, $\mathbf o$ and  $\mathbf{a}_k$; $Q_{\text{intra}}$, $Q$}
\Output{$\theta^j$, $\epsilon^j$, $\phi^j$ for $j \in \ALPHABET J$}

Update the action policy, broadcast policy and termination parameters as follows. For all $j \in \ALPHABET J$,
\begin{align*}
\theta^j &\leftarrow \theta^j + \alpha_{\theta^j} \frac{\partial \log{\pi^{o^j,\theta^j}}(a^j_k\,|\,s^j)}{\partial \theta^j} Q^j_{\text{intra}}(s^j, \omega^j, a^j_k, \mathrm{br}^j_k)\\
\epsilon^j &\leftarrow \epsilon^j + \alpha_{\epsilon^j} \frac{\partial \log{\pi^{\omega^j,\epsilon^j}}(\mathrm{br}^j_k\,|\,s^j)}{\partial \epsilon^j} Q^j_{\text{intra}}(s^j_k, \omega^j, a^j_k, \mathrm{br}^j_k)\\
 \phi^j &\leftarrow \phi^j + \alpha_{\phi^j} \frac{\partial \beta^{\omega^j,\phi^j}(s^{'j})}{\partial \phi^j}\Big(Q^j(s^{'j}_k, \omega^j)-\\
 & \hskip 12em \max_{\omega^{' j} \in \Omega^j} Q^j(s^{'j}_k, \omega^{'j}) \Big)
\end{align*}
If $\beta^{\omega^j,\phi^j}$ terminates in $s^{j'}$ then choose a new $o^j$ according to $\mu$ (softmax or $\varepsilon$-greedy)\;
\Return{$\theta^j$, $\epsilon^j$, $\phi^j$ for $j \in \ALPHABET J$}
}
\caption{Distributed Option Improvement (DOI)}\label{alg:DOI}
\end{algorithm2e}

\clearpage
\bibliographystyle{unsrt}
\bibliography{ref}

\end{document}